\documentclass{article}

\PassOptionsToPackage{numbers, compress}{natbib}
\usepackage[preprint]{neurips_2025}

\usepackage[utf8]{inputenc} 
\usepackage[T1]{fontenc}    
\usepackage{hyperref}       
\usepackage{url}            
\usepackage{booktabs}       
\usepackage{amsfonts}       
\usepackage{nicefrac}       
\usepackage{microtype}      
\usepackage{xcolor}         
\usepackage{graphicx}       
\usepackage{amsmath}
\usepackage{float}

\usepackage{listings}
\usepackage{xcolor}

\definecolor{codekw}{RGB}{0,0,180}      
\definecolor{codecom}{RGB}{0,128,0}     
\definecolor{codestr}{RGB}{163,21,21}   
\definecolor{codenum}{RGB}{128,128,128} 
\definecolor{codedef}{RGB}{111,0,138}   

\lstdefinestyle{mintedclean}{
  language=Python,
  basicstyle=\ttfamily\small,
  commentstyle=\color{codecom},
  keywordstyle=\color{codekw}\bfseries,
  stringstyle=\color{codestr},
  numberstyle=\tiny\color{codenum},
  identifierstyle=\color{black},
  emph={self,Tensor,nn,einsum,linear},
  emphstyle=\color{codedef},
  showstringspaces=false,
  breaklines=true,
  tabsize=4,
  keepspaces=true,
  columns=fullflexible,
  numbers=left,
  numbersep=6pt,
  upquote=true,
  frame=none,
  backgroundcolor=\color{white},
}

\title{Finding Manifolds With Bilinear Autoencoders}
\workshoptitle{Mechanistic Interpretability Workshop}

\DeclareMathOperator*{\argmax}{arg\,max}
\DeclareMathOperator{\Tr}{Tr}
\DeclareMathOperator{\Diag}{Diag}

\author{
  Thomas Dooms \\
  University of Antwerp\\
  Antwerp, Belgium \\
  \texttt{doomsthomas@gmail.com}
  \And
  Ward Gauderis \\
  Vrije Universiteit Brussel \\
  Brussels, Belgium \\
  \texttt{ward.gauderis@vub.be}
}

\begin{document}


\maketitle

\begin{abstract}
Sparse autoencoders are a standard tool for uncovering interpretable latent representations in neural networks. Yet, their interpretation depends on the inputs, making their isolated study incomplete.
Polynomials offer a solution; they serve as algebraic primitives that can be analysed without reference to input and can describe structures ranging from linear concepts to complicated manifolds.
This work uses bilinear autoencoders to efficiently decompose representations into quadratic polynomials. We discuss improvements that induce importance ordering, clustering, and activation sparsity. This is an initial step toward nonlinear yet analysable latents through their algebraic properties.
\end{abstract}
 
\begin{figure}[H]
  \centering
  \includegraphics[width=0.55\textwidth]{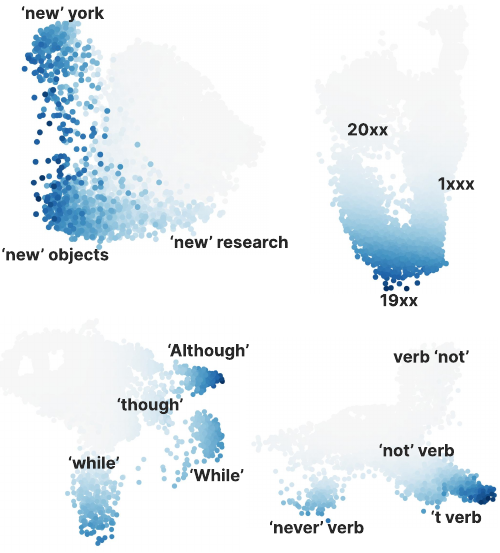}
  \caption{
  Conceptual subspaces extracted directly and solely from the weights of a bilinear autoencoder. To visualise the structure of these spaces, inputs are projected linearly onto it, with colour indicating activation strength. This reveals learned manifolds: the top-left organises nuances of 'new' into an off-centre triangle, while the top-right forms an off-centre circle for the first digits in a year. The bottom-left linearly clusters subordinating conjunctions, and the bottom-right captures verb negation. The top two are nonlinear manifolds, whereas the bottom two suggest linear concepts, possibly in superposition \citep{elhage2022toymodelssuperposition}. An interactive version is available at \url{https://mi-manifold.netlify.app/}.
  }
  \label{fig:manifold-examples}
\end{figure}

\addtocounter{footnote}{1}
\footnotetext{The code and models for this paper can be found at \url{https://github.com/tdooms/bae}.}

\section{Introduction}
\label{sec:intro}

Neural networks are highly nonlinear systems shaped by emergent phenomena rather than explicit mechanisms. Yet, human intuition is grounded in explicit structure and predictable composition. Mechanistic interpretability attempts to bridge this gap by identifying interpretable bases \citep{olah2022mechanistic,cunningham2023sparseautoencodershighlyinterpretable}, enabling latents to be (roughly) examined in isolation, despite being nonlinear \citep{ameisen2025circuit}. Yet, the core problem remains: these latents lack a compositional interpretation --- their isolated behaviour cannot be related to that of the whole \citep{coecke2021compositionalityit}. Extracted latents have implicit domains of applicability: regions in input or activation space within which they are meaningful. For instance, latent splitting and absorption \citep{chanin2025absorptionstudyingfeaturesplitting} split a latent's domain to increase sparsity while retaining its function. Consequently, their composition has a different meaning: instead of combining independent features, their domain is extended.
As a result, the isolated study of latents is of limited value, since their composition may skew their interpretation, even at tiny scales \citep{méloux2025everythingeverywhereoncemechanistic}. Compositional understanding is vital for interpretability, demanding that dependencies and interactions be explicit and computable \citep{tullCompositionalInterpretabilityXAI2024,gauderisCompositionalFrameworkInterpreting}.

One natural question is whether we can design or constrain nonlinear systems to remain compositionally interpretable \citep{tullCompositionalInterpretabilityXAI2024}. Polynomials offer a solution as they are nonlinear yet fully analysable through their coefficients, independent of their representation basis \citep{doomsCompositionalityUnlocksDeep24,pearce2025bilinearmlpsenableweightbased,gauderisCompositionalFrameworkInterpreting}. They also admit direct geometric interpretation: many familiar manifolds, such as hyperspheres, are described by polynomial equations. These properties make polynomials a promising foundation for studying nonlinear computation and representations, offering a glimpse into their learned geometries (\autoref{fig:manifold-examples}).

This paper investigates bilinear autoencoders \citep{gauderisCompositionalFrameworkInterpreting}, which efficiently decompose neural representations into polynomial latents. This enables the uncovering of manifolds in an automated manner. It also allows exact computation of metrics such as similarity and distance between or within autoencoders, without relying on input-dependent correlation analysis. In short, this work contributes the following:

\textbf{Architecture.} We give an introduction to bilinear autoencoders, outlining their conceptual motivation and an efficient implementation.

\textbf{Extensions.} We propose several new extensions that induce clustering, ordering and sparsity, avoiding post-hoc regularisation penalties.

\textbf{Analysis.} We use bilinear autoencoders to uncover interesting nonlinear manifolds and leverage their compositionality to quantify latent consistency.

\section{Bilinear autoencoders}
\label{sec:bae}

This manuscript assumes some familiarity with tensor notation and adopts the diagrammatic tensor notation from \citep{gauderisCompositionalFrameworkInterpreting} (for a short introduction, see \citep{TensorNetwork_Diagram_Notation}).
We denote scalars (0-order tensors) as $s$ (lower-case), vectors (1-order tensors) as $\mathbf{v}$ (bold lower-case), matrices (2-order tensors) as $M$ (upper-case) and higher-order tensors as $\mathbf{T}$ (bold upper-case). We represent indexed tensors as $\mathbf{T}_{ijk}$, and denote slicing along tensor dimensions as $\mathbf{T}_{m:n}$ to extract subtensors.

\subsection{Motivation and architecture} \label{sec:vanilla}

Autoencoders reconstruct their inputs through a bottleneck, forcing the model to capture structure in the data \citep{hinton2006reducingdimensionality,hindupurProjectingAssumptionsDuality2025}. They recently became a key tool in interpretability for extracting latents from neural representations \citep{cunningham2023sparseautoencodershighlyinterpretable, bricken2023monosemanticity}. In this setting, the goal is to reconstruct and relate the discovered structure to the original representations, imposing a trade-off between latent expressivity and analysis complexity.

At the simplest level, linear decompositions such as singular value decomposition (SVD) can be considered autoencoders. However, low-rank constraints are often too restrictive: neural representations typically span all dimensions, so purely rank-based methods can fail to capture much of their structure. A more expressive alternative is the (sparse) autoencoder, which adds a nonlinearity while keeping a linear decoder. This allows recovery of linearly separable latents that remain directly traceable to the original representations. Both approaches aim to reveal structures that are easily extractable from the representation, but make assumptions about how those features are represented \citep{hindupurProjectingAssumptionsDuality2025}
As a result, they can produce latents that reflect the imposed prior rather than the underlying structure itself \citep{engels2025languagemodelfeaturesonedimensionally,chanin2025absorptionstudyingfeaturesplitting}.

Bilinear autoencoders \citep{gauderisCompositionalFrameworkInterpreting} adapt sparse autoencoders to gated linear units \citep{shazeer2020gluvariantsimprovetransformer, dauphin2017languagemodelinggatedconvolutional}. GLUs apply two matrices to their input $\mathbf{x} \in \mathbb{R}^\text{In}$ and multiply the output element-wise $\odot$, acting as a dynamic gate:
\[
\mathrm{GLU}(\mathbf{x}) \;=\; D\big(\sigma(L \mathbf{x}) \odot R \mathbf{x}\big).
\]
The function $\sigma$ can be removed with negligible impact on accuracy \citep{pearce2025bilinearmlpsenableweightbased,shazeer2020gluvariantsimprovetransformer}. The resulting operation can represent second-order polynomials of its inputs while remaining linearly analysable in the `lifted' 
\textit{product space} of all pairwise multiplicative interactions $\mathbf{x} \otimes \mathbf{x}$ between elements of $\mathbf{x}$. A single encoded \textit{bilinear latent} $\mathbf{f}(\mathbf{x})_j$ ($j \in {1, \dots, \text{Lat}}$) is defined as.
\[
\mathbf{f}(\mathbf{x})_j \;:=\; \mathbf{x}^\top\mathbf{l}_j\mathbf{r}_j^\top \mathbf{x} \;=\; (\mathbf{l}_j^\top \otimes \mathbf{r}_j^\top)\,(\mathbf{x} \otimes \mathbf{x}) \;:=\; B_j X
\]
where $X = \mathbf{x} \otimes \mathbf{x}$ is the interaction vector (a 2-order tensor) and $B_j = (\mathbf{l}_j \otimes \mathbf{r}_j)^\top$ is a rank-1 matrix. Each bilinear latent thus corresponds to a rank-1 bilinear form. Stacking all $B_j$ into a 3-order tensor yields the encoder $\mathbf{B}$ of shape $(\text{Lat} \times \text{In} \times \text{In})$. When $\mathbf{B}$ is reshaped into a $(\text{Lat} \times \text{In}^2)$ matrix, we denote it as $B$.

The decoder is the encoder's transpose, trained to reconstruct the original product representation. Untying these weights has been observed to have a negligible impact on performance metrics. Hence, the bilinear autoencoder approximates $X$ as
\[
\hat{X} \;=\; B^\top B X
\]
Here, $\mathbf{f} := BX$ has fewer dimensions than $X = \mathbf{x} \otimes \mathbf{x}$ but more than $\mathbf{x}$, creating a (not necessarily sparse) bottleneck on the product space. 
This bottleneck corresponds to the amount of linearly extractable structure in the interaction matrix $X$, which is quadratic in $\mathbf{x}$.

\begin{figure}[H]
  \centering
  \includegraphics[width=0.7\textwidth]{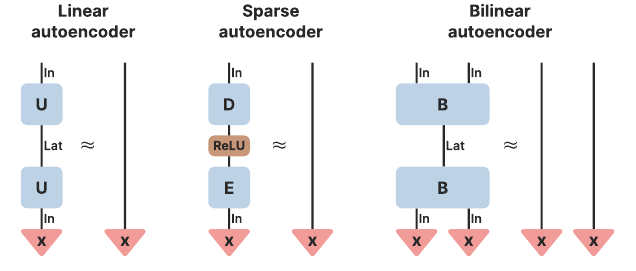}
  \caption{Most autoencoders reconstruct their inputs nonlinearly. Instead, bilinear autoencoders linearly reconstruct the product space $X$, which is quadratic in the input $\mathbf{x}$.}
  \label{fig:autoencoders}
\end{figure}

\subsection{Bilinear latents and their geometry}
\label{sec:latents}

\begin{figure}[H]
  \centering
  \includegraphics[width=0.35\textwidth]{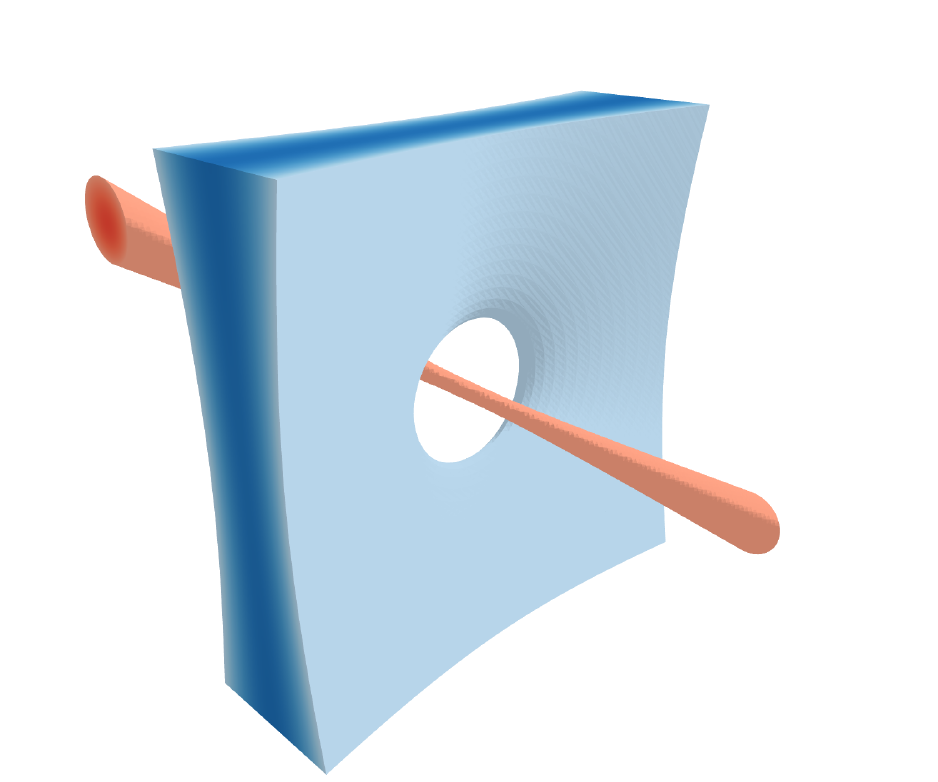}
  \caption{Coloured areas illustrate high activations $\mathbf{f}_i(\mathbf{x}) > 1$ for two bilinear latents $B_i$. Isolated latents can selectively activate on superposed directions (red). Composite latents can activate on more complicated manifolds, such as oriented cones and paraboloids (blue).}
  \label{fig:superposition}
\end{figure}

The primary motivation behind current autoencoders \citep{gao2024scalingevaluatingsparseautoencoders,rajamanoharan2024jumpingaheadimprovingreconstruction} is the linear representation hypothesis, claiming that most neural representations are encoded linearly \citep{park2024linearrepresentationhypothesisgeometry,cunningham2023sparseautoencodershighlyinterpretable}. However, recent research suggests some salient concepts span more complicated manifolds, which we would also like to capture \citep{modell2025originsrepresentationmanifoldslarge,engels2025languagemodelfeaturesonedimensionally}. Consequently, we want simple primitives which can be natively and meaningfully composed to encompass manifolds of varying complexity.
Polynomials fit this role: latents capture linear directions in product space, and their closed-form bilinear composition yields quadratic manifolds in input space (\autoref{fig:superposition}).

Single bilinear latents are similar to their ReLU-based counterparts in spirit; they activate strongly within cone-like regions. However, there are three subtle differences: bilinear latents are point symmetric across the origin, can have negative outputs, and do not have sharp activation boundaries. Despite the latter, latents still activate very selectively, as discussed in \autoref{app:zoom}.

Linear combinations of bilinear latents still have a closed form: they are represented by higher-rank bilinear forms. Each form defines an activation landscape, whose level sets manifest as quadratic surfaces. This enables combinations of latents to define intricate regions of high/low activation.

Bilinear latents' algebraic properties enable tractable analysis since they form a vector space; a single latent or any linear combination can be expressed as a bilinear form. This closure also yields basis independence, where rotations and mixtures of latents stay inside the same analysable vector space. This results in mathematical objects in which composite interactions can be made explicit and be manipulated, naturally extending interpretability from individual latents to their combinations \citep{gauderisCompositionalFrameworkInterpreting}.

\subsection{Efficient training with the kernel trick} \label{sec:kernel}

Evaluating bilinear autoencoders differs from ordinary ones since it reconstructs the quadratic space. While the inputs are naturally factored as $X = \mathbf{x} \otimes \mathbf{x}$, this is not true for the outputs $\hat{X} \neq \hat{\mathbf{x}} \otimes \hat{\mathbf{x}}$. Hence, naively computing the reconstruction loss $||\hat{X}-X||^2_F$ would require materialising this quadratic space. While this is feasible in small quantities, it becomes a significant hurdle for large-scale training. However, we can avoid this expensive computation by decomposing the sum of squares error (SSE; \autoref{app:setup}) into terms.

\begin{align}
\mathrm{SSE}(\hat{X}, X)
&= \sum_{i=1}^{\text{In}} (B^\top B X - X)_i^2
= X^\top B^\top B B^\top B X - 2 X^\top B^\top B X + X^\top X
\label{eq:vanilla-loss}
\end{align}

\begin{figure}[H]
  \centering
  \includegraphics[width=0.9\textwidth]{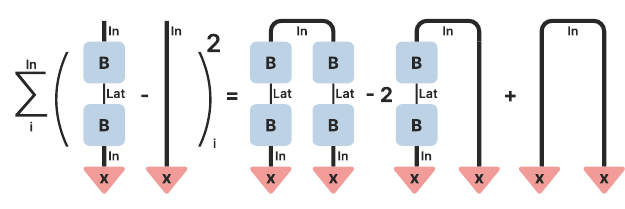}
  \caption{Diagrammatic formulation of \autoref{eq:vanilla-loss}. Lines indicate tensor contractions over that index, while unconnected lines represent unmodified open indices. Thick lines represent a quadratic space.}
  \label{fig:vanilla-loss}
\end{figure}

Rewriting the latent representation $BX$ as the vector $\mathbf{f}$, we get the following formula where ${B}{B}^\top$ is a matrix of shape ($\text{Lat} \times \text{Lat}$).
\[
\mathrm{SSE}(\hat{X}, X) \;=\; \mathbf{f}^\top\, ({B}{B}^\top)\, \mathbf{f} \;-\; 2 \mathbf{f}^\top \mathbf{f} \;+\; X^\top X
\]
This reframing avoids materialising the reconstruction. The latter two terms are inner products, and the first can be efficiently computed (\autoref{app:efficiency}). This is reminiscent of the "kernel trick" for polynomials, which simulates an intractable feature space by computing efficient inner products using a similarity kernel. In this case, the inner product is parameterised. We refer to ${B}{B}^\top$ as the kernel matrix of the bilinear autoencoder. 

In summary, evaluating bilinear autoencoders scales quadratically in time and linearly in memory with respect to the latent dimension. Despite the quadratic size of the kernel matrix, materialising it is not required. 
The term $\mathbf{f}^\top({B}{B}^\top)\mathbf{f}$ is efficiently computable through contraction ordering and tiling to scale well (similar to flash attention \citep{dao2022flashattentionfastmemoryefficientexact} and face-splitting products). See \autoref{app:efficiency} for details.

\section{Extensions} \label{sec:extensions}

Thus far, we have motivated and described a basic bilinear autoencoder. Here, we provide methods to efficiently induce additional structure into the learned latents. Specifically, we discuss activation sparsity, latent ordering and clustering. Note that these methods can apply to any autoencoder with a (multi-)linear decoder, but are not the focus of this present work.

\subsection{Avoiding dead latents with scale-invariant sparsity} \label{sec:sparsity}

Activation sparsity is a good prior for finding monosemantic latents that correspond to a single concept. This sparsity is often enforced by minimising sparsity-inducing norms, like $L_{0 < p \le 1}$, of the activations. Yet, this penalty punishes absolute magnitudes, driving all activations to zero in its optimum. This trivial shrinkage of absolute scale is misaligned with our goal of promoting \emph{relative} prominence. This can destabilise training by encouraging dead latents. We avoid this by using a scale-invariant measure, removing pressure toward shrinkage. Specifically, we use the Hoyer density measure \citep{hoyer}, which smoothly measures the relative density of a latent $\mathbf{f}$.
\begin{equation}
\mathrm{Hoyer}_{\text{density}}(\mathbf{f}) \;=\; \frac{\frac{\|\mathbf{f}\|_1}{\|\mathbf{f}\|_2} - 1}{\sqrt{\text{Lat}}-1}
\label{eq:hoyer}
\end{equation}
This measure ranges from 0 for one-hot vectors to 1 for perfectly uniform ones. Vectors with Gaussian entries score $0.8$ in expectation, while a $k$-sparse vector yields at most $k/\text{Lat}$. We compute this value for each latent across all samples (batch $\times$ sequence). This batch-wise minimisation exerts dual pressure for latent variables to activate sharply but avoid collapse, producing selective yet robust latent representations \citep{bussmann2024batchtopksparseautoencoders}. By adding the Hoyer density as a penalty with coefficient $\alpha$ to the SSE loss, we obtain the following objective (where $\textbf{f}_j$ are latent activations over a batch):
\begin{equation}
\mathcal{L}(\mathbf{x}) \;=\; 
\mathrm{SSE}(\hat{X}, X) \;+\;
\alpha \sum^{\text{Lat}}_j \mathrm{Hoyer}_{\text{density}}(\mathbf{f}_j),
\label{eq:loss}
\end{equation}

\subsection{Ordering latents by importance through cumulative reconstruction} \label{sec:order}

Another desideratum for autoencoders is that they admit an explicit ordering with respect to feature importance, much like SVD. This allows the autoencoder to be truncated in (near-)optimal way to retain reconstruction quality. Currently, existing methods order through stochastic sampling or retraining on residual errors \citep{bussmann2025learningmultilevelfeaturesmatryoshka}. Ideally, we would like to impose total ordering during training. This is achieved by using a (weighted) cumulative reconstruction loss, promoting all prefix sets to achieve good reconstruction.

\begin{equation}
\mathrm{SSE}_{\text{cum}}(\hat{X}, X)
\;=\; \sum_{i = 1}^{\text{In}} \sum_{k = 1}^{\text{Lat}} w_k\, \big({B}^\top ({B}X)_{1:k} - X\big)_i^2
\;=\; \mathbf{f}^\top\big(({B}{B}^\top) \odot (M^\top \Diag(\textbf{w}) M) \big) \mathbf{f} \;-\; \cdots
\label{eq:cum}
\end{equation}

\begin{figure}[H]
  \centering
  \includegraphics[width=0.9\textwidth]{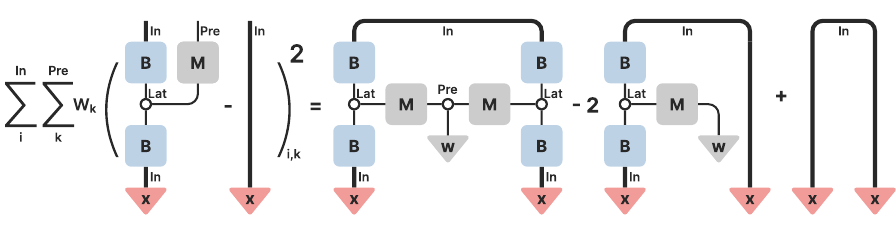}
  \caption{Diagrammatic formulation of the ordered loss. Here, $M$ is an upper triangular mask matrix with columns $\mathbf{m}_k$ and $\mathbf{w}$ are the weights of each cumulative reconstruction. We use a uniform distribution for $\mathbf{w}$ to make all prefix reconstructions ($\text{Pre}$) equally important. This encourages the latents to contribute to the reconstruction in an approximately arithmetic progression of importance.}
  \label{fig:ordered-loss}
\end{figure}

For each output element, we sum over all prefix reconstruction errors. This is achieved by multiplying the latents with a mask $\mathbf{m}_k = [\mathbf{1}_k, \mathbf{0}_{d-k}]^\top$.  By collecting $\mathbf{m}_k$ as columns into a matrix, forming an upper-triangular mask matrix $M$, and collecting $\sqrt{w_k}$ into a vector $\mathbf{w}$, we can write
\[
\sum_{k = 1}^{\text{Lat}} w_k \big({B}^\top (\mathbf{f} \odot \mathbf{m}_k)\big)_i^2
\;=\;
\big({B}^\top \mathbf{f} \otimes (M^\top \Diag(\textbf{w}) M)\big)_{i}^2
\]
Naively computing this suffers from the issues described in \autoref{sec:kernel}: materialising these products is expensive. Hence, we perform the same trick and recast the ordered loss as a kernel matrix. Specifically, instead of ${B}{B}^\top$ we obtain $({B}{B}^\top) \odot (M^\top \Diag(\textbf{w}) M)$ which can be efficiently computed. The other terms left out in \autoref{eq:cum} remain relatively unchanged, becoming a weighted inner product. Empirically, this yields autoencoders that reconstruct well with fewer latents (\autoref{app:ordering}).

\subsection{Mixing latents through a linear bottleneck} \label{sec:mix}

Not all latents are one-dimensionally linear; some concepts span intricate multi-dimensional manifolds \citep{engels2025languagemodelfeaturesonedimensionally}. The challenge is finding latent combinations that likely contain meaningful structure. We introduce a bottleneck within the latent space, using a down-projection $D$ into a lower-dimensional space, producing mixture latents whose bilinear form can attain higher rank.
\[
\hat{X} \;=\; {B}^\top D^\top D\, {B}\, X
\]
This is reminiscent of the setup of \citep{elhage2022toymodelssuperposition}:
imposing the bottleneck encourages the model to identify (anti-)correlations among latents to improve reconstruction. In this way, the bottleneck highlights interacting multi-dimensional linear spaces that are likely to contain more complex manifolds.

\subsection{Combining extensions} \label{sec:combination}



The extensions introduced above are modular and can be combined to produce autoencoder setups with different behaviour. In this paper, we apply ordering and sparsity to the latent basis. Alternatively, one could apply sparsity within the bottleneck from \autoref{sec:mix}, promoting sparse latent mixtures.

Combining these extensions may require attention to how penalties interact. In particular, ordering via cumulative reconstruction amplifies the contribution of early latents to the loss, making them more important for reconstruction. To preserve balance, other penalties such as activation sparsity should be scaled accordingly across the ordered basis.

\begin{equation}
\mathcal{L}(\mathbf{x}) \;=\;
\mathrm{SSE}(\hat{X}, X) \;+\;
\alpha \sum^\text{Lat}_k \sum^\text{Lat}_j (\text{Lat} - k) \mathrm{Hoyer}_{\text{density}}(\mathbf{\textbf{f}_j})
\label{eq:cumulative_penalty}
\end{equation}

\section{Experiments} \label{sec:experiments}

This section provides experimental evidence for the utility of bilinear autoencoders. We discuss their reconstruction and sparsity trade-off, how they help find manifolds, and their similarity across hyperparameters. We experimented with the following four different flavours of bilinear autoencoders.

\begin{itemize}
    \item \texttt{vanilla}: contains only $L$ and $R$ matrices and uses Hoyer sparsity (\autoref{sec:sparsity}).
    \item \texttt{ordered}: adds a penalty to order latents by importance (\autoref{sec:order}).
    \item \texttt{mixed}: adds an additional $D$ matrix to cluster/mix latents (\autoref{sec:mix}).
    \item \texttt{combined}: a combination of mixed and ordered (\autoref{sec:combination}). 
\end{itemize}

\subsection{Measuring sparsity and reconstruction}

Bilinear autoencoders differ from nonlinear variants in that they cannot learn hard cutoffs. Nevertheless, they achieve strong reconstruction, indicating that neural representations are well-separable within the product space. 
Due to their different reconstruction objective, direct comparison to sparse autoencoders is subtle. \autoref{app:comparison} shows that bilinear autoencoders can reconstruct the product space more effectively than sparse autoencoders.

\begin{figure}[H]
  \centering
  \includegraphics[width=0.55\textwidth]{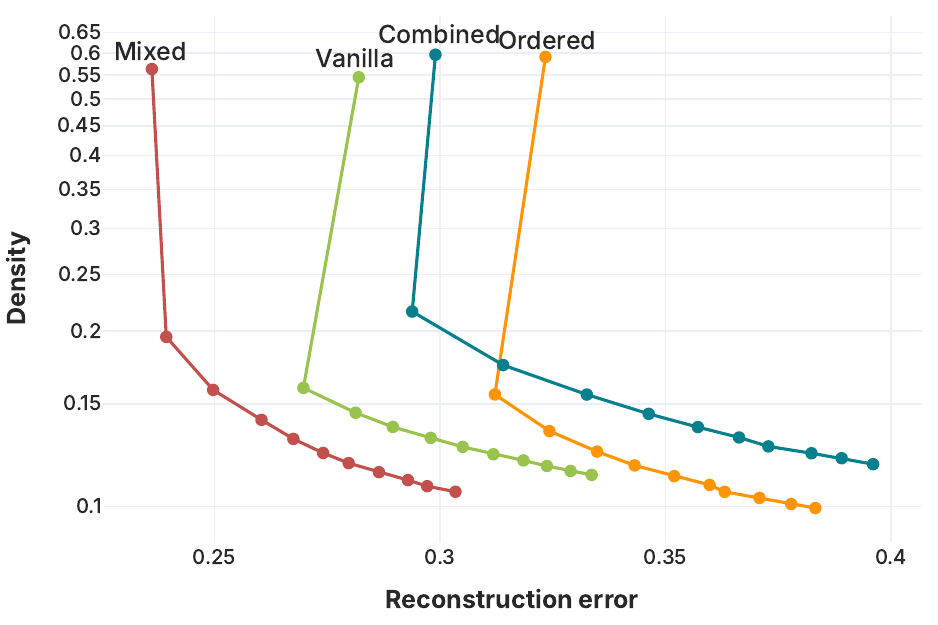}
  \caption{Reconstruction error (\autoref{eq:vanilla-loss}) vs. activation density (\autoref{eq:hoyer}) across variants and sparsity penalties between $\alpha = 0$ (top) and $\alpha = 1$ (bottom) on layer 18 of Qwen3-0.6B. Lower error and density are better: activations can be strongly sparsified with minimal reconstruction trade-offs. The reconstruction is normalised between $0$ and $1$ (\autoref{app:setup}).}
  \label{fig:pareto}
\end{figure}

Density–reconstruction analysis (\autoref{fig:pareto}) shows that activation density (\autoref{eq:hoyer}) can be freely lowered from $0.6$ to $0.2$ (by increasing $\alpha$ from $0$ to $0.1$) at no cost to reconstruction, beyond which errors increase. While latent mixtures improve this trade-off, imposing an ordering degrades it. Intriguingly, density penalties sometimes improve reconstruction, suggesting that the model is learning naturally sparse features.

\begin{figure}[H]
  \centering
  \includegraphics[width=0.48\textwidth]{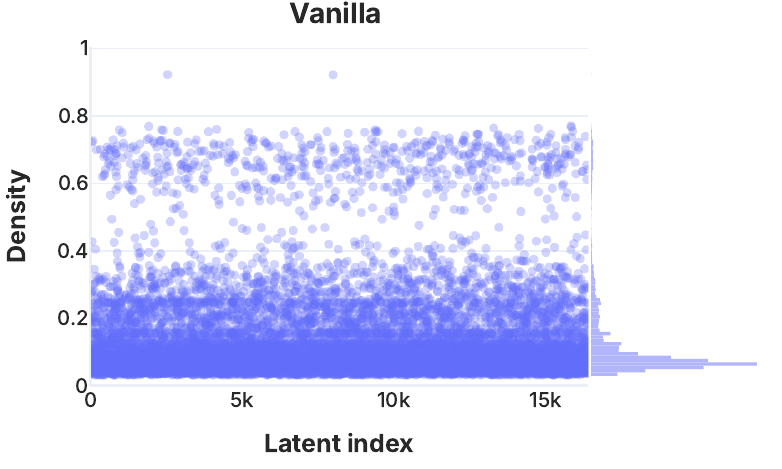}
  \includegraphics[width=0.48\textwidth]{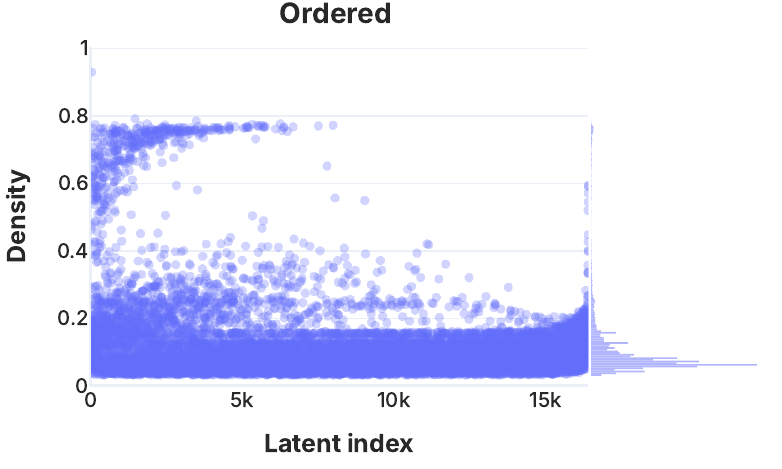}
  \caption{The Hoyer density of latents for the vanilla and ordered autoencoders (both $\alpha = 1$). This reveals an apparent dichotomy where about $2\%$ of latents activate densely despite sparsification.}
  \label{fig:sparsity-distribution}
\end{figure}

The distribution of activation sparsities reveals further structure, especially when ordered (\autoref{fig:sparsity-distribution}). The \texttt{vanilla} and \texttt{ordered} variants exhibit one prominent peak of highly sparse latents and a smaller peak of dense latents. The ordering pushes the dense latents to the front, emphasising their importance \citep{sun2025densesaelatentsfeatures}. 
Notably, this effect is independent of sparsity regularisation, as the penalties are adjusted (\autoref{sec:combination}).

\subsection{Discovering manifolds through latent interactions} \label{sec:manifolds}

Some features in language models live neatly along linear directions. Others occupy curved or folded manifolds that such linear views cannot capture \citep{modell2025originsrepresentationmanifoldslarge}. To identify these manifolds, we train a \texttt{mixed} autoencoder (\autoref{sec:mix}) whose reconstruction passes through a constrained latent basis.
\[
\mathbf{f}^\top {D}^\top {D} \mathbf{f} \;\approx\; I
\]
The matrix ${D}^\top {D}$ encodes how latents co-behave. Clusters in this matrix signal latent groups that interact during reconstruction and are likely to contain multi-dimensional latent structure \citep{elhage2022toymodelssuperposition}. We quantify these strongly interacting latents by counting the number of `high' elements in their rows. This is measured using the row's densities after some sharp function, in this case a fourth power: $\text{Hoyer}_\text{density}((D^\top D)_i^4)$. Similarly-sized outliers remain relevant while everything else is pushed to zero. The density measure is high if the original vector contains mostly similar-sized entries, while it will be low if it has a single outlier. We then consider the top 10 entries in our candidate cluster and combine them into a (weighted) composite latent.

Since this composite latent is a bilinear form, we can use its eigendecomposition to find the most salient $3\mathrm{D}$ subspace in the original input space. Next, we define the activation strength of a composite latent as the $L_2$ norm of its individual activations. We then project the $64\mathrm{k}$ highest activating inputs (out of $256\mathrm{k}$) onto the extracted subspace. The result is a fully linear slice of the input space, so any visible structure must be present in the original representation. In contrast to methods like PCA, this approach reproducibly reveals manifolds purely from the autoencoder’s weights, without relying on input-dependent statistics.

The visualisations (\autoref{fig:manifold-examples}) reveal diverse geometries: smooth quadratic surfaces (top left and right), sharply separated clusters (bottom left), and linear directions (bottom right). We inspected about 50 candidates and picked the most interesting ones. Qualitatively, most concepts are linearly represented; their visualisation yields a large cloud of near-zero activations with one or a few directions sticking out with high activation.
However, roughly half showed additional nonlinear structure. Such cases illustrate how the latent space can contain nonlinear manifolds beyond isolated directions. See \autoref{app:random} for a visualisation of random latents.

\subsection{Computing exact reconstruction similarity between autoencoders} \label{sec:similarity}

How consistent are autoencoders? In most settings, the answer comes from partial evidence: a few encoder/decoder directions align \citep{paulo2025sparseautoencoderstraineddata}, and reconstruction correlations are roughly the same. Bilinear autoencoders can give an exact weight-based answer to this question, without referencing a data distribution. Since they can be represented by a (matricised) 4-order tensor ${A} = {B}^\top {B}$, we can compare their Frobenius similarity directly.

\begin{equation}
\mathrm{Similarity}_{\mathrm{Frobenius}}({A}, {A}') \;=\; 1 - \frac{\|{A} - {A}'\|_F^2}{\|{A}\|_F^2 + \|{A}'\|_F^2} = \frac{2 \Tr({A}^T{A}')}{\|{A}\|^2_F + \|{A}'\|^2_F}
\label{eq:frobenius}
\end{equation}

\begin{figure}[H]
  \centering
  \includegraphics[width=\textwidth]{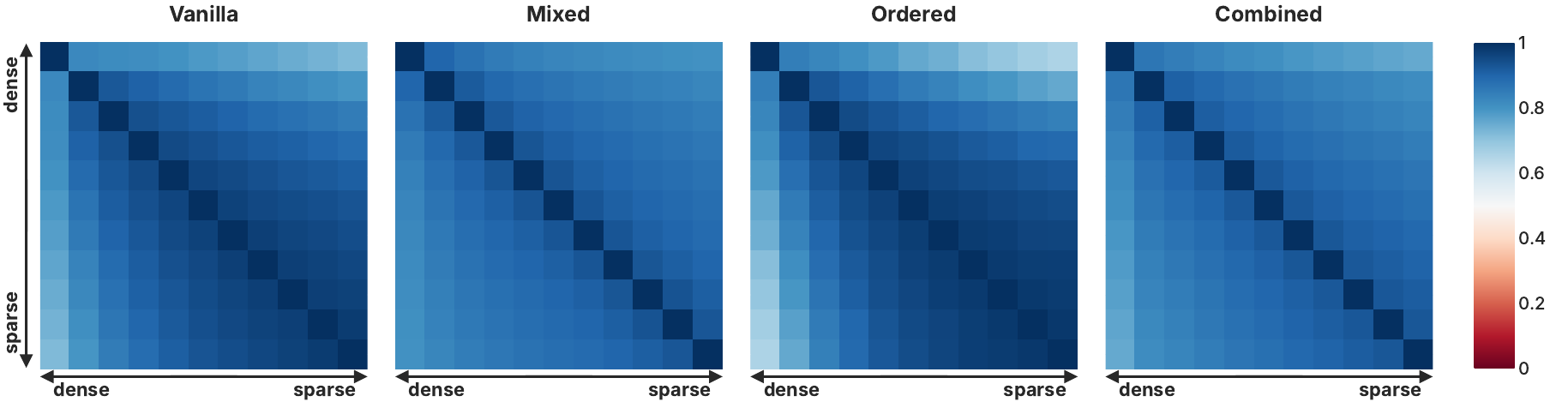}
  \caption{Frobenius similarity between bilinear autoencoders with density penalties for $\alpha \in [0, 1]$ (from \autoref{fig:pareto}). This shows they learn comparable reconstructions across hyperparameters.}
  \label{fig:frobenius}
\end{figure}

Bilinear autoencoders with comparable sparsity are near-identical ($98\%$ in \texttt{vanilla} and \texttt{ordered} \autoref{fig:frobenius}). For instance, a \texttt{vanilla} autoencoder attains over $99\%$ similarity when varying seeds, demonstrating the consistency of the learning process. However, this similarity drops significantly when comparing the densest and sparsest instances. This shows that sparsity penalties not only impact the latents but also the behaviour of the whole.

The Frobenius similarity above is global and basis-independent: it compares the exact reconstruction between autoencoders \footnote{This similarity can be adapted to account for ordering by including $M^\top M$ into the equation. This can be interpreted as a cumulative similarity but is not explored in this paper.}. This is often not sufficiently precise since we may be interested in latent consistency: how similarly each latent is represented. Linearity is helpful again since it is possible to construct a matrix that maps how latents are represented on another autoencoder's basis. We explore this idea further and provide empirical results in \autoref{app:similarity}.

\section{Conclusion} \label{sec:conclusion}
\paragraph{Summary.} This paper uses bilinear autoencoders to decompose representations into polynomial latents. We argue that polynomial latents are suitable primitives for interpretability, as they can be analysed without inputs and help reveal composite geometries. We propose several extensions to induce latents with additional structure: sparsity, importance ordering, and clustering. Lastly, we show that our theory translates to practice by enabling the automated discovery of manifolds.

\paragraph{Future work.} This paper studies bilinear autoencoders and highlights their advantages across tasks. Much work remains to rigourise results and explore the whole state space of possible applications:
\begin{itemize}
\item \textbf{Performing causal analysis and steering.} Using the interpretable manifolds as a basis for (automated) causal interventions to verify and nonlinearly steer model behaviour.
\item \textbf{Enabling multi-layer interpretability.} Linearly mapping between the latent bases of stacked autoencoders to trace how features are composed through a deep network.
\item \textbf{Investigating composite manifolds.} Designing sparsity penalties that align interacting latents to represent composite concepts, moving beyond the sparsity-induced single-latent alignment explored in this work.
\item \textbf{Incorporating geometric biases.} Integrating architectural priors, such as bias terms, to restrict latents to specific (known) geometries.
\item \textbf{Scaling to practical applications.}  Applying these autoencoders to larger models and latent spaces to discover new circuits, debug model failures or verify safety properties.
\end{itemize}

\paragraph{Implications.} This work moves beyond (near-)linear decompositions and promotes polynomials as algebraic primitives, as their computation and complexity can be analysed in a closed form. This can facilitate quantifying their description length, which is often intractable or ill-defined for other nonlinear operations without making assumptions \citep{ayonrinde2024interpretabilitycompressionreconsideringsae}. This approach can help alleviate reliance on interpretable bases in favour of studying composition and interactions \citep{gauderisCompositionalFrameworkInterpreting}.

\begin{ack}
We thank Michael Pearce, Jose Oramas, and Logan Riggs for fruitful discussions and useful feedback on the draft. This research received funding from the Flemish Government under the "Onderzoeksprogramma Artificiële Intelligentie (AI) Vlaanderen” programme.
\end{ack}

\section{Contributions}
TD and WG jointly developed the theory on sparse bilinear autoencoders. TD conducted experiments, developed the extensions, wrote the code, and composed the initial manuscript. WG provided theoretical guidance to support formalisation and collaborated on manuscript development.

\bibliographystyle{plainnat}
\bibliography{refs}


\appendix
\section{Experimental setup} \label{app:setup}

We use the same hyperparameters across all discussed autoencoders. Since we only performed a few ablations, this setup is very likely suboptimal.

\paragraph{Data.} We performed all experiments on layer 18 of \texttt{Qwen3-0.6B-Base} \citep{qwen3}, a small but competent model. This lets us sweep over reasonably sized autoencoders and fully train on a tight compute budget. Specifically, the \texttt{vanilla} autoencoder trains in about 10 minutes on an RTX 4080. We successfully replicated these experiments on other models of varying sizes and ages (\autoref{app:anecdata}).

\paragraph{Optimiser.} We use the Muon optimiser \citep{jordan2024muon}, which we have found to significantly outperform Adam. This depends strongly on the setup, but it's roughly between $2\times$ and $10\times$ faster convergence. We use a trapezoidal learning rate schedule, cooling down to $0$ over the last half of training. Finally, we use a linear warmup for $\alpha$ to prioritise reconstruction over sparsity near the start.

\paragraph{Architecture.} We have found orthogonal initialisation (\autoref{app:code}) to perform better than Xavier initialisation. Orthogonal initialisation favours sparser solutions at a minute reconstruction cost. We use an expansion factor of $16$, a mixing expansion of $2$ (\autoref{sec:mix}), and $\alpha=0.1$ by default.

\paragraph{Objective.} We divide inputs with their $L_2$ norm and use the sum of squares error (SSE) over the more common RMS and MSE combination. This is done for convenience, removing some clutter from equations.
Since the autoencoder is linear, this normalisation does not affect the optimisation objective beyond ensuring equivalently-scaled inputs, improving stability.

\begin{figure}[H]
  \centering
  \begin{minipage}{0.31\textwidth}
    \centering
    \begin{tabular}{@{}ll@{}}
      \textbf{Batch size} & $32 \times 2$ \\
      \textbf{Sequence length} & $256$ \\
      \textbf{Steps} & $2^{10}$ \\
      \textbf{Total tokens} & $\approx 8\mathrm{M}$ \\
      \textbf{Shuffling} & No \\
    \end{tabular}
  \end{minipage}\hfill
  \begin{minipage}{0.31\textwidth}
    \centering
    \begin{tabular}{@{}ll@{}}
      \textbf{Optimiser} & Muon \\
      \textbf{Momentum} & No \\
      \textbf{Learning rate} & $0.01$ \\
      \textbf{Schedule} & trapezoid \\
      \textbf{$\alpha$ warmup} & $256$ steps \\
    \end{tabular}
  \end{minipage}\hfill
  \begin{minipage}{0.34\textwidth}
    \centering
    \begin{tabular}{@{}ll@{}}
      \textbf{Initialisation} & Orthogonal \\
      \textbf{Model dimension} & $1024$ \\
      \textbf{Latent dimension} & $16 \times 1024$ \\
      \textbf{Mix dimension} & $2 \times 1024$ \\
      \textbf{Default $\alpha$} & $0.1$ \\
    \end{tabular}
  \end{minipage}
  \caption{Hyperparameters related to data, the optimiser, and the architecture, respectively.}
  \label{fig:hyperparameters}
\end{figure}

\section{Efficient evaluation of the loss} \label{app:efficiency}

In \autoref{sec:kernel}, we claim that using the kernel trick to train bilinear autoencoders is not expensive. Here, we discuss this claim in more detail. We first demonstrate that computing $\mathbf{B}^\top \mathbf{B}$ is efficient and then show that it can be calculated with an efficient contraction order and splitting to avoid memory allocation.

In short, the kernel matrix of a bilinear autoencoder can be efficiently evaluated by recasting it as an element-wise multiplication of two matrices. 

\begin{equation}
{B}{B}^\top
= (LL^\top) \odot (RR^\top)
\label{eq:kernel}
\end{equation}

\begin{figure}[H]
  \centering
  \includegraphics[width=0.6\textwidth]{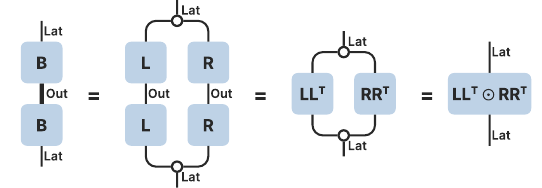}
  \caption{Diagrammatic equation on how to compute a bilinear autoencoder's kernel.}
  \label{fig:kernel}
\end{figure}

This perhaps also provides intuition on the ordering kernel matrix, which is also efficiently computable since it is simply an additional element-wise multiplication. The extra bottleneck in the \texttt{mixed} autoencoder only adds a matrix to the contraction, shrinking the kernel matrix.

While this trick reduces computational complexity, the kernel matrix is quadratic in the number of latents in memory, which can still be a bottleneck. However, it can be evaluated by splitting the $\text{Lat}\otimes\text{Lat}$ space into $n$ blocks $s_1 \oplus s_2 \oplus \dots \oplus s_n$, which avoids full materialisation.
\[
\mathbf{f}^\top BB^\top \mathbf{f} = \sum_{s_i, s_j} \mathbf{f}_{s_i}\, (BB^\top)_{s_i s_j}\, \mathbf{f}_{s_j}
\]
This can be additionally improved since only the symmetric part of $BB^\top$ contributes. Hence, we need only sample the lower or upper triangular part, speeding up computation by a factor $2$.

This technique extends to all proposed autoencoder variants. In the \texttt{mixed} case, the down projection is one extra step in the kernel evaluation. In the \texttt{ordered} case, the mask matrix can be element-wise multiplied with each block.

\section{Autoencoder pseudo code} \label{app:code}

We provide distilled pseudo code for the initialisation and forward pass of the vanilla autoencoder. The intuition for why this works is provided in \autoref{app:efficiency}. The full code can be found at \url{https://github.com/tdooms/bae}.






\begin{minipage}{\linewidth}
\begin{lstlisting}[style=mintedclean]
def __init__(self, config):
    self.left = nn.init.orthogonal(config.d_features, config.d_model)
    self.right = nn.init.orthogonal(config.d_features, config.d_model)

def forward(self, acts: Tensor['batch seq dims'], alpha: float):
    # Normalise the inputs and compute the features
    acts = acts * acts.square().sum(-1, keepdim=True).rsqrt()
    features = linear(acts, self.left) * linear(acts, self.right)

    # Compute the density and regularisation term
    density = hoyer(features, dim=(0, 1)).mean()

    # Compute constituent parts of the reconstruction error
    kernel = (self.left @ self.left.T) * (self.right @ self.right.T)
    recons = einsum(f, f, kernel, "... h1, ... h2, h1 h2 -> ...")
    cross = f.square().sum(-1)

    # Compute the error and loss
    error = (recons - 2 * cross + 1.0).mean()
    loss = error + alpha * density
    return loss, dict(error=error, density=density)
\end{lstlisting}
\end{minipage}

\section{Comparison of reconstruction error} \label{app:comparison}

\begin{figure}[H]
  \centering
  \includegraphics[width=\textwidth]{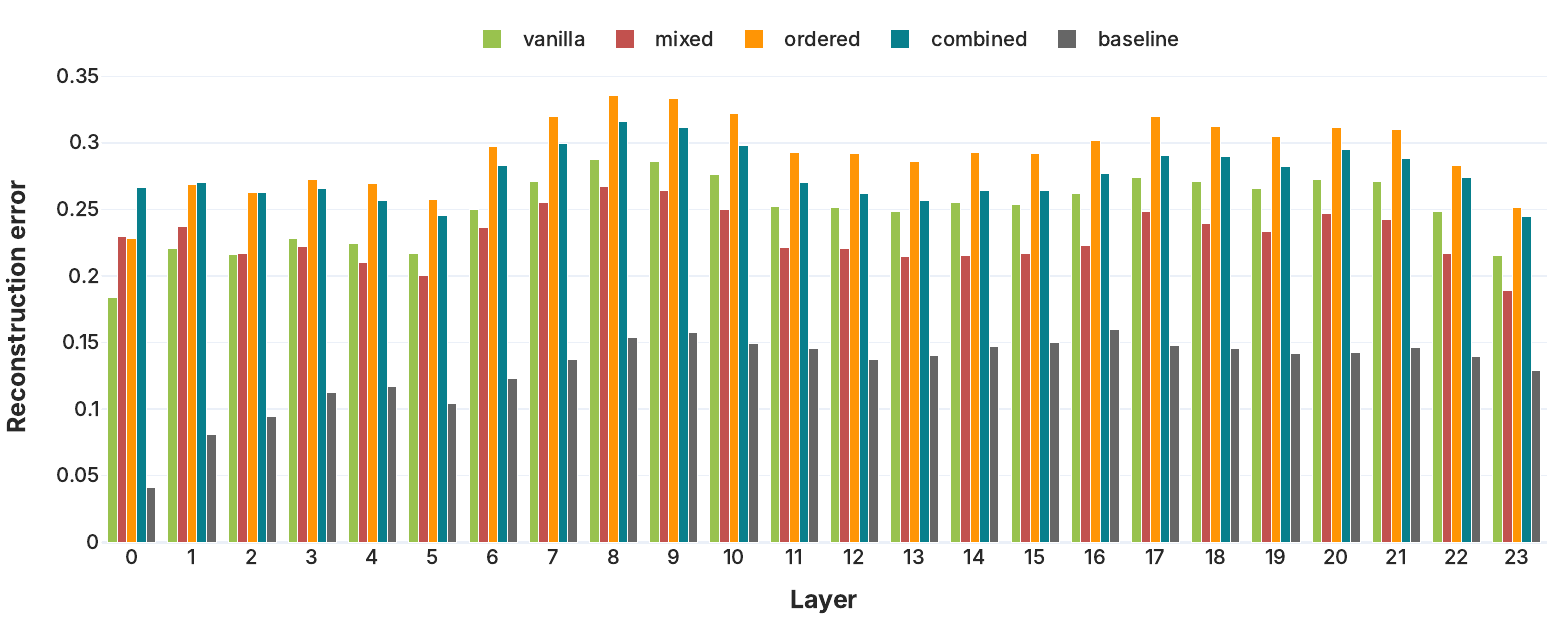}
  \caption{Reconstruction error across layers and autoencoder variants. The baseline comparison is unfair and should be doubled (\autoref{eq:quadratic-recons}). Bilinear autoencoders often outperform the baseline in product space reconstruction.}
  \label{fig:layer-recons}
\end{figure}

We compare the reconstructions of all bilinear autoencoders across layers with a standard TopK ($k = 50$) autoencoder \citep{gao2024scalingevaluatingsparseautoencoders} baseline trained using the same setup (\autoref{fig:layer-recons}). This baseline comparison is unfair since the bilinear variants reconstruct the product input space rather than the inputs themselves. We work out the discrepancy, starting from the TopK reconstruction error:
\[
s \;=\; \|\mathbf{x} - \hat{\mathbf{x}}\|^2 \;=\; \|\mathbf{x}\|^2 - 2\mathbf{x}^\top \hat{\mathbf{x}} + \|\hat{\mathbf{x}}\|^2.
\]
Due to the input normalisation (\autoref{app:setup}), we have $\|x\|^2 = 1$ and $\|\hat{x}\|\approx 1$, so we can approximate the quadratic reconstruction $S$ given the regular reconstruction as follows.

\begin{align}
S &= \|\mathbf{x} \otimes \mathbf{x} - \hat{\mathbf{x}} \otimes\hat{\mathbf{x}}\|^2 \nonumber \\
  &= 1 - 2 (\mathbf{x}^\top \hat{\mathbf{x}})^2 + \|\hat{\mathbf{x}}\|^4 \nonumber \\
  &= \tfrac{1}{2} (1 - \|\hat{\mathbf{x}}\|^2)^2 + s(1 + \|\hat{\mathbf{x}}\|^2) - \tfrac{1}{2} s^2 \nonumber \\
  &\approx 2s - \tfrac{1}{2} s^2
\label{eq:quadratic-recons}
\end{align}

For fair comparison, the baseline error should roughly be doubled ($s^2 \approx 0$ when small). Consequently, \autoref{fig:layer-recons} shows that bilinear autoencoders achieve better reconstruction than the baseline across most model layers and variants in the product space. 
This suggests that bilinear autoencoders can leverage decodable structure in interactions between input elements to improve reconstruction.

Another interesting observation is that \texttt{mixed} autoencoders outperform their \texttt{vanilla} counterparts everywhere except the first model layers. The \texttt{mixed} autoencoder adds a bottleneck, indicating that latents are better described as composites than independently. The fact that this doesn't happen in the first few layers may indicate these composite structures have not yet formed, but this is purely speculation.

\section{Anecdata and miscellaneous ablations} \label{app:anecdata}

\paragraph{Convergence.} All autoencoders are trained on $16\text{M}$ tokens, which takes a few minutes. This is sufficient to get highly interpretable features and interesting manifolds. While we lack the computational resources to train on billions of tokens, we train a $32\times$ \texttt{mixed} autoencoder on $256\text{M}$ tokens. Training metrics decrease steadily throughout training, down to an $\text{SSE}$ of 0.14. This makes us optimistic that bilinear autoencoders can scale properly with compute.

\paragraph{Ordering.} When the learning rate decreases, ordered autoencoders exhibit interesting learning dynamics near the end of training. The reconstruction error increases, sparsity remains the same, and loss decreases. This likely indicates that its features are being reshuffled and perhaps hints that the training process struggles with this task under certain circumstances and may require cyclic learning rates to alternate shuffling and learning.

\paragraph{Models.} \texttt{Qwen3} models use the modern SwiGLU \citep{qwen3}, which aligns well with bilinear encoders. We verify whether bilinear autoencoders work well on older models like GPT2-small \citep{radford2019language} and Pythia-160M \citep{biderman2023pythia} and, more importantly, whether they produce interpretable features. The answer is positive even though the reconstruction error is about $20\%$ higher on equivalent setups. More broadly, we find that more competitive models of similar size yield lower reconstruction errors.

\paragraph{Gates.} Despite removing many appealing properties, we experimented with using nonlinearities in encoders to reflect an actual GLU. While this (slightly) reduces sparsity, it also yielded worse reconstructions, resulting in overall higher loss. Furthermore, we also experimented with applying the activation function after the element-wise multiplication. Using TopK required setting $k > 500$ to get competitive reconstructions with other bilinear autoencoders. In both cases, we uncoupled the encoder and decoder since their roles are no longer identical.

\paragraph{Bounded.} The imposed latent bottleneck by reconstructing the squared space is bounded by the rank of $\mathbf{x} \otimes \mathbf{x}$. In theory, increasing the latent space beyond the size of the product space cannot reduce the reconstruction error any further, unless additional assumptions such as sparsity are imposed \citep{leICAReconstructionCost}. This insight can be used to predict the best achievable reconstruction error given random inputs for a given size. We can use this as a baseline to discern whether scaling the latent amount simply reconstructs noise or still finds structure. We find autoencoders up to an expansion factor $64$ outperformed this random baseline, but did not scale this experiment further.

\paragraph{Muon.} All experiments use the Muon optimiser \citep{jordan2024muon} by default. It's hard to overstate the improvement over Adam; it often reduces training time by an order of magnitude. Furthermore, it's much more tolerant to changes in hyperparameters and changes in training setup. This was not the case for Adam in our experience.

\section{Reconstruction with ordered latents} \label{app:ordering}

\begin{figure}[H]
  \centering
  \includegraphics[width=0.6\textwidth]{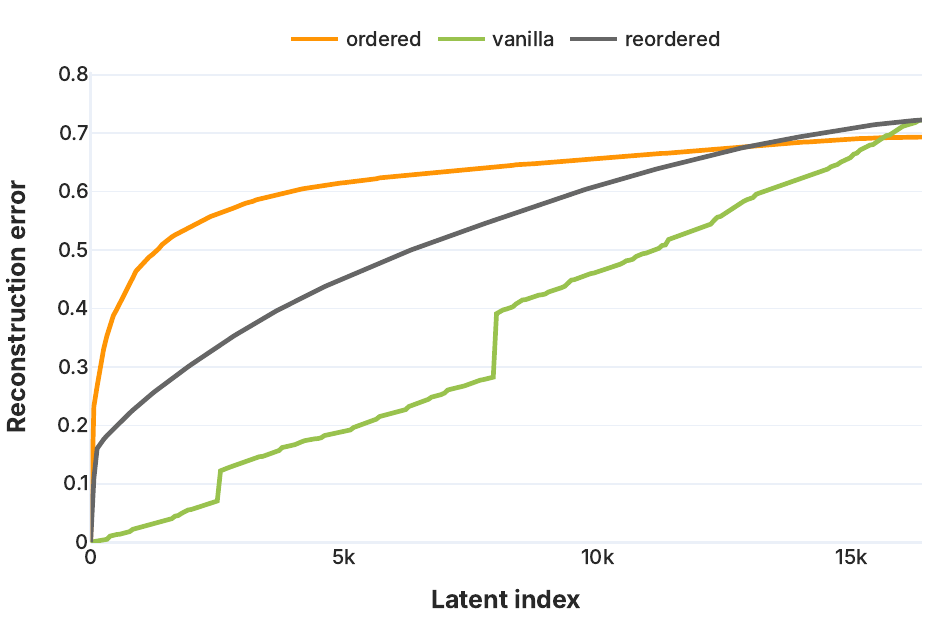}
  \caption{Reconstruction errors for all prefix subsets of latents on two autoencoder variants. The `reordered' baseline uses the best greedy post-hoc ordering for the vanilla reconstruction for comparison. This is computed by recursively adding the latent which improves reconstruction the most. }
  \label{fig:ordered-recons}
\end{figure}

This section empirically examines the effect of imposing an ordering on latents. Our results demonstrate that ordering influences the final reconstruction (\autoref{fig:layer-recons}), highlighting a trade-off that deserves closer analysis. Here, we show that ordered autoencoders achieve lower reconstruction errors significantly faster than their baseline (\autoref{fig:ordered-recons}).

The trained \texttt{ordered} and \texttt{combined} autoencoders use cumulative reconstruction errors. Hence, the first latent's reconstruction is often orders of magnitude more important than the last, which may be too `strong'. Instead, one could consider a gentler schedule that only puts mild penalties on later latents. However, we have not yet experimented with how this impacts the latents' behaviour.

\section{Zooming in on a latent} \label{app:zoom}

\begin{figure}[H]
  \centering
  \includegraphics[width=0.6\textwidth]{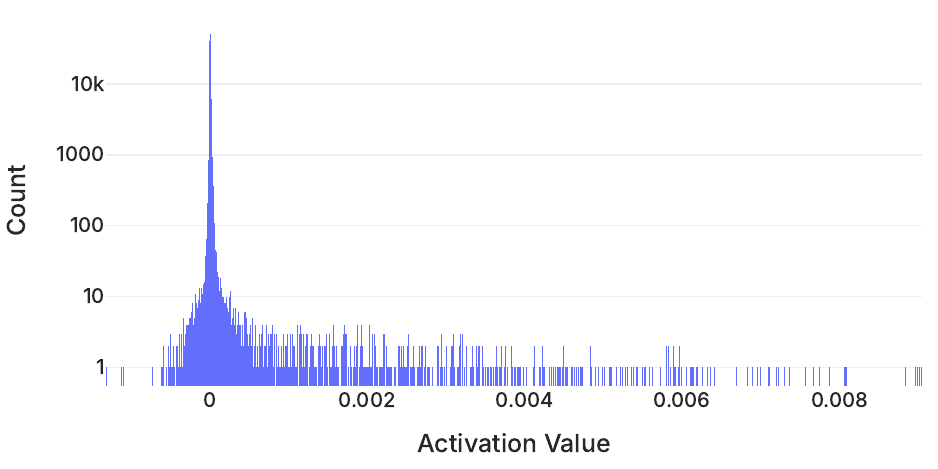}
  \caption{Histogram of $128\mathrm{k}$ activations for the first latent of \texttt{vanilla} ($\alpha = 1$) autoencoder. Due to the logarithmic y-axis, the peak spans multiple orders of magnitude.}
  \label{fig:activation-distribution}
\end{figure}

This section discusses the first latent in the \texttt{vanilla} autoencoder without sparsity penalty (shown as the most dense model in \autoref{fig:pareto}). Its activations correspond to possessive pronouns (``their'', ``her'', ``its''). Due to the lack of a sparsity-inducing activation function and symmetry about the origin, there is a question of how selectively it activates. We sample $128\mathrm{k}$ activations and plot their magnitude, showing that $99\%$ are smaller than $2 \cdot 10^{-5}$ (\autoref{fig:activation-distribution}). Qualitative analysis shows this is a representative distribution for other latents; sometimes the tail is denser or longer, but this rough shape is always present.

\section{Random Manifolds} \label{app:random}

\begin{figure}[H]
  \centering
  \includegraphics[width=0.7\textwidth]{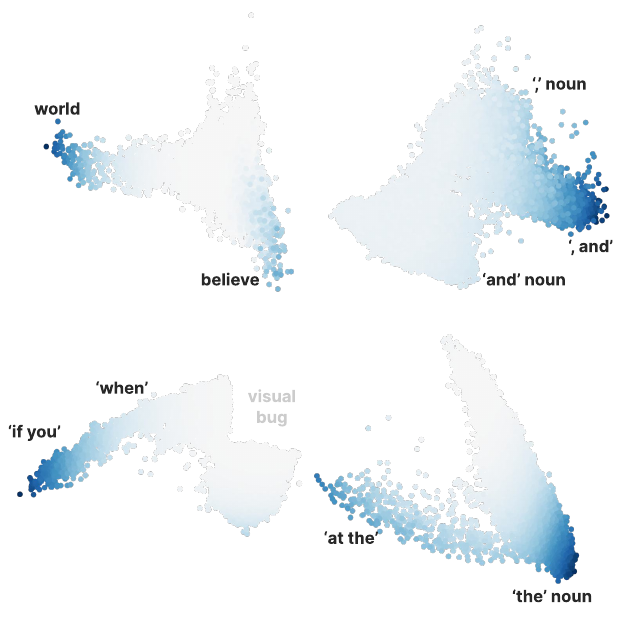}
  \caption{
  Subspace visualisations of the first 4 latents to demonstrate non cherry-picked manifolds.
  The top left has two separate concepts: one side activates on words like 'world', 'global' and 'international' while the other on 'believe', 'think', 'feel'. These representations likely share a similar subspace without deeper semantic meaning. The top right contains conjunctions where the strongly activating space activates on the bigram ', and' while either side activates on only one of these tokens followed by a noun. The bottom left reveals a single direction that activates on conditional clauses, most strongly on 'if'. Only the top activating samples are shown so if there is a long tail of small activations, this causes an illusory gap in the visualisation that would contain very low activations. Finally, the bottom-right contains an off-centre linear direction that contains the bigram 'at the'. Nearby samples activate on 'the' followed by mostly nouns.
  }
  \label{fig:random-manifolds}
\end{figure}

The manifolds in \autoref{fig:manifold-examples} are cherry-picked through a combination of quantitative and qualitative filtering. About 1 hour of time was spent in this process. This raises a natural question of what most other latents look like, which is answered in \autoref{fig:random-manifolds}.

\section{Similarity between latent bases} \label{app:similarity}

\autoref{sec:similarity} showed that the learned structure is remarkably similar across autoencoders with varying hyperparameters. Yet, there is a question regarding local identifiability: do they learn similar latents across runs? This question can again be answered exactly for bilinear autoencoders.

This analysis is ordinarily based on the cosine similarity of decoder directions. Yet, this is incomplete as a latent's behaviour depends on its output direction (in the decoder) and its input direction (in the encoder). Furthermore, tiny changes in either encoder or decoder weights can strongly impact behaviour due to the activation functions, which are often ignored. Since bilinear latents can be studied independently of the inputs, their nonlinear behaviour can be analysed precisely.

A latent $i$ is fully described by the matrix $B_i = \mathbf{l}_i \otimes \mathbf{r}_i$. Contracting two encoders like this ${B}^\top {B}'$ creates a (latent $\times$ latent) matrix which captures all pairwise similarities. When discussing identifiability, we seek this matrix to be as close to a permutation $P$ as possible. In words, \autoref{eq:permutation-similarity} measures how much of the norm of ${B}^\top {B}'$ the best permutation $P$ can recover.

\begin{equation}
\mathrm{Similarity}_{\mathrm{Perm}}({B}, {B}') \;=\; \frac{\|{B'} P {B}^\top\|_F}{\|{B}^\top {B}'\|_F} \quad \text{where} \quad P = \argmax_{\text{permutation }P} \text{Tr}({B}' P {B}^\top)
\label{eq:permutation-similarity}
\end{equation}

\begin{figure}[H]
  \centering
  \includegraphics[width=0.4\textwidth]{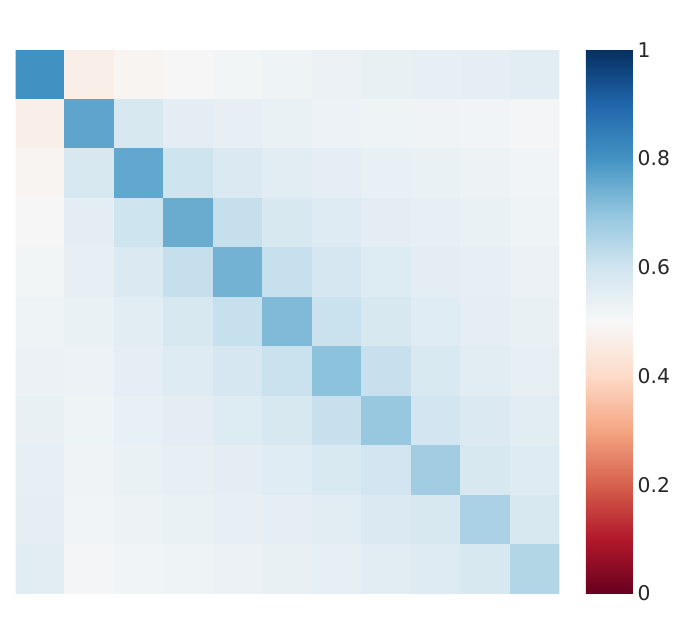}
  \caption{Permutation similarities (\autoref{eq:permutation-similarity}) across \texttt{vanilla} autoencoder with varying sparsity penalties $\alpha \in [0, 1]$.}
  \label{fig:permutation}
\end{figure}

\autoref{fig:permutation} show the resulting similarities for the \texttt{vanilla} autoencoder with varying sparsities.

The diagonal describes the self-similarity of an autoencoder and is not guaranteed to be $1$ since decoders are the encoder's transpose (not inverse). It indicates the amount of interference (nonorthogonality) between latents and can be interpreted as how similar latents are after a `round-trip' through an autoencoder. For the \texttt{vanilla} autoencoders from \autoref{fig:pareto}, this value lies between $0.8$ for densely activating autoencoders and $0.65$ for their sparse counterparts.

The off-diagonal entries contain cross-similarities of two autoencoders. Note that this similarity is rather strict since it measures the similarity of the latents' bilinear forms, which is more sensitive than the cosine similarity of the vectors alone. This value is relatively constant and lies between $0.5$ and $0.6$.

Lastly, we believe that consistency of the whole (\autoref{fig:frobenius}) matters more than latent identifiability (\autoref{fig:permutation}) for bilinear autoencoders. In linear models, we rarely insist on aligning bases because our tools are not basis-dependent. Unlike nonlinear autoencoders, which fix their privileged latent basis using activation functions, bilinear autoencoders behave linearly in the product space, enabling rotation and combination of latents without leaving the bilinear form (\autoref{sec:latents}). Hence, analyses remain equivalent regardless of the used basis.

\end{document}